%% file: Large_VAR.tex
\documentclass[a4paper,12pt]{article}
\usepackage{amsfonts}
\usepackage[english]{babel}
\usepackage{graphicx}
\usepackage[usenames]{color}
\usepackage{colordvi}
\usepackage{amssymb}
\usepackage{amsmath}
\usepackage[round]{natbib}
\usepackage{multirow}
\usepackage{multicol}
\usepackage{ifthen}
\usepackage{ulem}
\usepackage{xcolor,eso-pic}
\usepackage{colordef}

{\quad ${\Box}$ \newline}
\def\proofname{Proof}

\def\remarkname{REMARK}
\newtheorem{theorem}{THEOREM}[section]

\newtheorem{assumption}{ASSUMPTION}[section]

\newtheorem{lemma}{LEMMA}[section]

\begin{document}
\input{defs}
\title{Large Vector Auto Regressions}
\author{Song Song\thanks{University of California, Berkeley. Email: songsong@stat.berkeley.edu}, Peter J. Bickel \thanks{University of California, Berkeley. Email: bickel@stat.berkeley.edu}}
\maketitle
\begin{abstract}
One popular approach for nonstructural economic and financial forecasting is to include a large number of economic and financial variables, which has been shown to lead to significant improvements for forecasting, for example, by the dynamic factor models. A challenging issue is to determine which variables and (their) lags are relevant, especially when there is a mixture of serial correlation (temporal dynamics), high dimensional (spatial) dependence structure and moderate sample size (relative to dimensionality and lags). To this end, an \textit{integrated} solution that addresses these three challenges simultaneously is appealing. We study the large vector auto regressions here with three types of estimates. We treat each variable's own lags different from other variables' lags, distinguish various lags over time, and is able to select the variables and lags simultaneously. We first show the consequences of using Lasso type estimate directly for time series without considering the temporal dependence. In contrast, our proposed method can still produce an estimate as efficient as an \textit{oracle} under such scenarios. The tuning parameters are chosen via a data driven ``rolling scheme'' method to optimize the forecasting performance. A macroeconomic and financial forecasting problem is considered to illustrate its superiority over existing estimators.

{\it Keywords}: Time Series, Vector Auto Regression, Regularization, Lasso, Group Lasso, Oracle estimator

{\it JEL classification}: C13, C14, C32, E30, E40, G10
\end{abstract}

\section{Introduction}\label{intro}
Macroeconomic forecasting is one of the central tasks in Economics. Broadly speaking, there are two approaches, structural and nonstructural forecasting. Structural forecasting, which aligns itself with economic theory, and
hence rises and falls with that, recedes following the decline of Keynesian theory. In
recent years, new dynamic stochastic general equilibrium theory has been
developed, and structural macroeconomic forecasting is poised for resurgence. Nonstructural
forecasting, in contrast, attempts to exploit the reduced-form correlations in observed
macroeconomic time series, has little reliance on economic theory, has always been working well and
continues to be improved. Various univariate and multivariate time series analyzing techniques have been proposed, e.g. the auto regression (AR), moving average (MA), autoregressive moving average (ARMA),  generalized autoregressive conditional heteroskedasticity (GARCH), vector auto regression (VAR) models among many others. A very challenging issue for this nonstructural approach is to determine which variables and (their) lags are relevant. If we omit some ``important'' variables by mistake, it potentially creates an omitted variable bias with adverse consequences for both structural analysis and forecasting. For example, \cite{Ch:Ei:Ev:99} points out that the positive reaction of prices in response to a monetary tightening, the so-called \textit{price puzzle}, is an artefact resulting from the omission of forward-looking variables, such as the commodity price index. Recently, \cite{RePEc:jae:japmet:v:25:y:2010:i:1:p:71-92} shows that, when using the cross-sectional dimension related shrinkage, the forecasting performance of small monetary vector auto regression can be improved by adding additional macroeconomic variables and sectoral information. To illustrate this, we consider an example of interest rate forecasting. Nowadays people primarily use univariate or multivariate time series models, e.g. the Vasicek, CIR, Jump-Diffusion, Regime-Switching, and time-varying coefficients models, all of which are mostly based on the information from the interest rate time series itself. However, in practice, the central bank (Fed) bases their decisions of interest rate adjustment (as a monetary policy instrument) heavily on the national macroeconomic situation by taking many macro and financial measures into account. Bringing in this additional spatial (over the space of variables instead of from a geographic point of view; also used in future for convenience) information will therefore help improve its forecasting performance. Another example about the interactions between macroeconomics and finance comes from modeling credit defaults by also using macroeconomic information, since variation in aggregate default rates over time presumably reflects changes in general economic conditions also. \cite{fi:fr:li:06} find credit events are significantly affected by macroeconomic factors. Not only macroeconomics could affect finance, finance could also affect macroeconomics. For example, the economic crisis typically starts from the stock market crash. All of these call for an \textit{integrated} analysis of macroeconomics and finance. Thus recently there has been a growing trend of using large panel macroeconomic and financial time series for forecasting, impulse response study and structural analysis, \cite{RePEc:tpr:restat:v:82:y:2000:i:4:p:540-554}, \cite{RePEc:bes:jnlasa:v:97:y:2002:m:december:p:1167-1179}, \cite{RePEc:bes:jnlbes:v:20:y:2002:i:2:p:147-62}, also seen at \cite{RePEc:bes:jnlasa:v:100:y:2005:p:830-840}, \cite{sto:wat:05}, \cite{RePEc:nbr:nberch:6670}, and \cite{RePEc:jae:japmet:v:25:y:2010:i:1:p:71-92} for latest advancements.

Besides its presence in empirical macroeconomics, high dimensional data, where information often scatters through a large number of interrelated time series, is also attracting increasing attention in many other fields of economics and finance. In neuro-economics and behavioral finance, one uses high dimensional functional magnetic resonance imaging data (fMRI) to analyze the brain's response to certain risk related stimuli as well as identifying its activation area, \cite{wor:02} and \cite{am:s2:pm:hh:wh:10}. In quantitative finance, one studies the dynamics of the implied volatility surface for risk management, calibration and pricing purposes, \cite{fen:hae:mam:07}. Other examples and research fields for very large dimensional time series include mortality analysis, \cite{lee:car:92}; bond portfolio risk management or derivative pricing, \cite{nel:sig:87} and \cite{die:li:06}; international economy (many countries); industrial economy (many firms); quantitative finance (many assets) analysis among many others.

On the methodology side, if people still use either low dimensional (multivariate) time series techniques on a few subjectively (or from some background knowledge) selected variables or high dimensional ``static'' methods which are initially designed for independent data, they might either disregard potentially relevant information (temporal dynamics and spatial dependence) to produce suboptimal forecasts, or bring in additional risk. Examples include the already mentioned prize puzzle and interest rate forecasting problems. The more scattered and dynamic the information is, the severer this loss becomes. This modeling becomes more challenging under the situation that macroeconomic data we typically deal with has only low frequencies, e.g. monthly or yearly. For example, the popularly used dataset introduced by \cite{sto:wat:05a} contains $131$ monthly macro indicators covering a broad range of categories including income, industrial production, capacity, employment and unemployment, consumer prices, producer prices, wages, housing starts, inventories and orders, stock prices, interest rates for different maturities, exchange rates and money
aggregates and so on. The time span is from January $1959$ to December $2003$ (so $T=540$). In summary, we can see that the challenge of modeling high dimensional time series, especially the macroeconomic ones, comes from a mixture of serial correlation (temporal dynamics), high dimensional (spatial) dependence structure and moderate sample size (relative to dimensionality and lags). To this end, an \textit{integrated} solution addressing these three challenges simultaneously is appealing.

To circumvent this problem, dynamic factor models have been considered to be quite successful recently in the analysis of large panels of time series data, \cite{RePEc:tpr:restat:v:82:y:2000:i:4:p:540-554}, \cite{RePEc:bes:jnlasa:v:97:y:2002:m:december:p:1167-1179}, \cite{RePEc:bes:jnlbes:v:20:y:2002:i:2:p:147-62}, also seen at \cite{RePEc:bes:jnlasa:v:100:y:2005:p:830-840}, \cite{RePEc:nbr:nberch:6670}, \cite{RePEc:bes:jnlasa:v:104:i:485:y:2009:p:284-298} and \cite{s2:wh:jr:10} (nonstationary case). They rely on the assumption that the bulk of dynamics interrelations within a large dataset can be explained and represented by a few common factors (low dimensional time series). Less general models in the literature include \textit{static} factor models proposed by \cite{RePEc:bes:jnlasa:v:97:y:2002:m:december:p:1167-1179}, \cite{RePEc:bes:jnlbes:v:20:y:2002:i:2:p:147-62} and \textit{exact} factor model suggested by \cite{RePEc:fip:fedmwp:55} and \cite{ge:77}.

Compared with the well studied dynamic factor models through the use of dynamic principal component analysis, the vector auto regressive (VAR) models have several natural advantages. For example, compared with the dynamic factor models' typical $2$-step estimation procedure: dimension reduction first and low dimensional time series modeling, the VAR approach is able to model the high dimensional time series in one step, which may lead to greater efficiency. It also allows variable-to-variable relationship (impulse response) analysis and facilitates corresponding interpretation, which is not feasible in the factor modeling setup since the variables are ``represented'' by the corresponding factors.
Historically, the VAR models are not appropriate for analyzing high dimensional time series because they involve the estimation of too many ($J^2P$, where $J$ is the dimensionality and $P$ is the number of lags) parameters. Thus they are primarily implemented on relatively low dimensional situations, e.g. the Baysian VARs (BVAR) by \cite{do:li;si:84} or still through the idea of factor modeling, e.g. the factor-augmented VAR (FAVAR) by \cite{RePEc:tpr:qjecon:v:120:y:2005:i:1:p:387-422}. However, based on recent advances in variable selection, shrinkage and regularization theory from \cite{ti96}, \cite{RePEc:bes:jnlasa:v:101:y:2006:p:1418-1429} and \cite{RePEc:bla:jorssb:v:68:y:2006:i:1:p:49-67}, large unrestricted vector auto regression becomes an alternative for the analysis of large dynamic systems. Therefore, the VAR framework can also be applied to empirical problems that require the analysis of more than a handful of time series. \cite{DeMol2008318} and \cite{RePEc:jae:japmet:v:25:y:2010:i:1:p:71-92} proceed that from the Bayesian point of view. \cite{ch:pe:07} consider the case that $P=1$ and both $J$ and $T$ are large through some ``neighboring'' procedure, which can be viewed as a special case of the ``segmentized grouping'' as we study here (details in Subsection \ref{segg}). In the univariate case ($J=1$), \cite{RePEc:bla:jorssb:v:69:y:2007:i:1:p:63-78} studies the regression coefficient and autoregressive order shrinkage and selection via the lasso when $P$ is large and $T$ is small (relative to $P$).

In this article, we will study the large vector auto regressions when $J, P \rightarrow \infty$ and $T$ is moderate (relative to $JP$). Comparing to prior works in (large) vector auto regressions, the novelty of this article lies in the following perspectives. {\textit{First}}, from the variable selection and regularization point of view, the theoretical properties of many existing methods have been established under the independent scenario, which is rarely met in practice and contradicts the original time series setup (if used directly). Disregarding the serial correlation in variable selection and regularization can be dangerous in the sense that various risk bounds in fact depend on the degree of time dependence, as we will illustrate later. We propose a new methodology to address this serial correlation (time dependence) issue together with high dimensionality and moderate sample size, which enables us to obtain the consistency of variable selection even under the dependent scenario, i.e. to reveal the equilibrium among them. {\textit{Second}}, our method is able to do variable selection and lag selection simultaneously. In previous literature, variable selection is usually carried out first, and then the corresponding estimate's performances w.r.t. different number of lags are compared through some information criteria to select the ``optimal'' number of lags. By doing so, we neglect the ``interaction'' between variable selection and lag selection. Additionally, when the number of the lag's candidates to be searched over is large, it is also computationally inefficient, due to the cost of the repeated variable selection procedures. {\textit{Third}}, we differentiate the variable of interest's own lags (abbreviated as \textit{own lags} afterwards) from the ones of other variables (abbreviated as \textit{others' lags} afterwards). Their relative weights are also allowed to be varied when predicting different variables, while in other literature, they are assumed to stay the same. This is due to the fact that the dynamic of some variables is driven by itself, while for a different variable, it might be driven by the dynamics of others. When we include a vast number of macroeconomic and financial time series, assuming the same weight seems to be too restrictive. {\textit{Fourth}}, our method is based on a more computationally efficient approach, which mostly uses the existing packages, e.g. the LARS (least angle regression) package, developed by \cite{citeulike:3284841}, while most other works in the literature go through the Bayesian approach that requires the choice of priors.

The rest of the article is organized as follows. In the next section, we present the main ingredients of the large vector autoregressive model (Large VAR) together with several corresponding estimation procedures and comparisons among them. The estimates' properties are presented in Section \ref{asymptotic}. In Section \ref{application}, the method is applied to the motivating  macroeconomic forecasting problem, which shows that it outperforms some major existing method. Section \ref{discussion} contains concluding remarks with discussions about relaxing some assumptions. All technical proofs are sketched in the appendix.

\section{The Large VAR Model and Its Estimation}\label{metho}
In this section, we introduce the model with three different estimates first, then discuss the data driven choice of hyperparameters to optimize the forecasting performance, provide a numerical algorithm, and finally summarize comparisons among these three estimates.

\subsection{The Model}
Assume that the high dimensional time series $\{Y_{tj}\}_{t=1, j=1}^{T \;\;\;\; J}$ is generated from
\begin{eqnarray}
Y_t^\T &=& Y_{t-1}^\T B_1+ \ldots + Y_{t-P}^\T B_P + U_t^\T \label{eq:main0}\\
\underbrace{\left(
  \begin{array}{l}
    Y_T^\T \\
    Y_{T-1}^\T\\
    \ldots \\
    \end{array}
\right)}_{T\times J}&=&
\underbrace{\left(
  \begin{array}{llll}
    Y_{T-1}^\T & Y_{T-2}^\T & \ldots & Y_{T-P}^\T \\
    Y_{T-2}^\T & Y_{T-3}^\T & \ldots & Y_{T-1-P}^\T \\
    \ldots & \ldots & \ldots & \ldots \\
  \end{array}
\right)}_{T \times JP}
\underbrace{\left(
  \begin{array}{l}
    B_1 \\
    B_2\\
    \ldots \\
    \end{array}
\right)}_{JP \times J}+\underbrace{\left(
  \begin{array}{l}
    U_T^\T \\
    U_{T-1}^\T\\
    \ldots \\
    \end{array}
\right)}_{T \times J}\nonumber \\
Y &=&X B  +U, \hfill \qquad \textrm{(compact form)} \label{eq:main}
\end{eqnarray}
where
\begin{itemize}
\item $Y=(Y_T^\T, \ldots, Y_1^\T)^\T$ with $Y_t^\T=(Y_{t1}, \ldots, Y_{tJ})$;
\item $X=(X_T^\T, \ldots, X_1^\T)^\T$ (the lags of $Y$) with $X_t=(Y_{t-1}^\T, \ldots, Y_{t-P}^\T)^\T$;
\item $B_1, \ldots, B_P$ are $J \times J$ autoregressive matrices, where $P$ is the number of lags, $B=(B_1, \ldots, B_P)^\T$ is the $JP \times J$ matrix containing all coefficients $\{B_{pij}\}_{p=1, i=1, j=1}^{P \;\;\;\; J  \;\;\;\; J}$, and $B_{\cdot \cdot j}, B_{p \cdot j}, B_{\cdot i \cdot}, B_{p i \cdot}$ is the $j$th column of $B$ and $B_p$, $i$th row of $B$ and $B_p$ respectively;
\item $U=(U_T^\T, \ldots, U_1^\T)^\T$, where $U_t$ is a $J$-dimensional noise and independent of $X_t$.
\end{itemize}
All $Y_t, X_t$ and $U_t$ are assumed to have mean zero. The $J \times J$ covariance matrix of $U_t$, $\Cov(U_t)$, is assumed to be independent of $t$. Here we assume $\Cov(U_t)$ to be diagonal, say $I_{J*J}$. In our case, it is justified by the fact that the variables in the panel we will consider for estimation are standardized and demeaned. Similar assumption is also carried out in \cite{DeMol2008318}. The relaxation allowing nonzero off-diagonal entries is discussed in Section \ref{discussion}.

\begin{figure}[h]
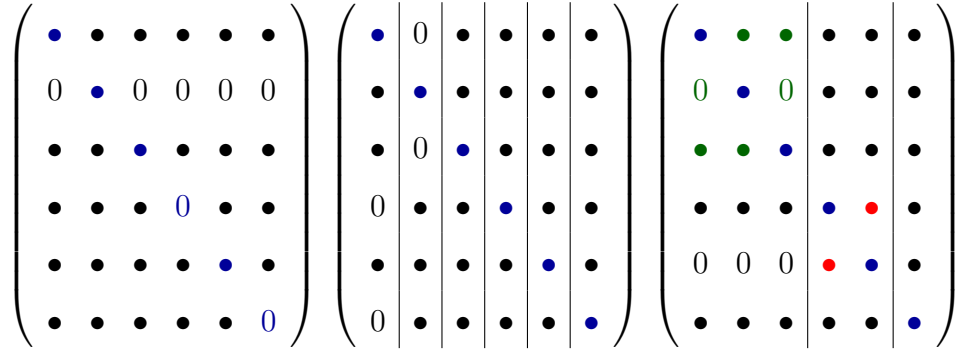

\centering    $\left(
  \begin{array}{cccccc}
{\color{iseblue}\bullet} & \bullet & \bullet & \bullet & \bullet & \bullet \\
    0 & {\color{iseblue}\bullet} & 0 & 0 & 0 & 0 \\
        \bullet & \bullet & {\color{iseblue}\bullet} & \bullet & \bullet & \bullet \\
            \bullet & \bullet & \bullet & {\color{iseblue}0} & \bullet & \bullet \\
                \bullet & \bullet & \bullet & \bullet & {\color{iseblue}\bullet} & \bullet \\
                    \bullet & \bullet & \bullet & \bullet & \bullet & {\color{iseblue}0}
  \end{array}
\right)$
    $\left(
  \begin{array}{c|c|c|c|c|c}
{\color{iseblue}\bullet} & 0 & \bullet & \bullet & \bullet & \bullet \\
    \bullet & {\color{iseblue}\bullet} & \bullet & \bullet & \bullet & \bullet \\
        \bullet & 0 & {\color{iseblue}\bullet} & \bullet & \bullet & \bullet \\
            0 & \bullet & \bullet & {\color{iseblue}\bullet} & \bullet & \bullet \\
                \bullet & \bullet & \bullet & \bullet & {\color{iseblue}\bullet} & \bullet \\
                    0 & \bullet & \bullet & \bullet & \bullet & {\color{iseblue}\bullet} \\
  \end{array}
\right)$
$\left(
  \begin{array}{ccc|cc|c}
{\color{iseblue}\bullet} & {\color{isegreen}\bullet} & {\color{isegreen}\bullet} & \bullet & \bullet & \bullet \\
{\color{isegreen}0} & {\color{iseblue}\bullet} & {\color{isegreen}0} & \bullet & \bullet & \bullet \\
        {\color{isegreen}\bullet} & {\color{isegreen}\bullet} & {\color{iseblue}\bullet} & \bullet & \bullet & \bullet \\
            \bullet & \bullet & \bullet & {\color{iseblue}\bullet} & {\color{isered}\bullet} & \bullet \\
                0 & 0 & 0 & {\color{isered}\bullet} & {\color{iseblue}\bullet} & \bullet \\
                    \bullet & \bullet & \bullet & \bullet & \bullet & {\color{iseblue}\bullet}
  \end{array}
\right)$\\
  \caption{Illustration of three different types of estimates.}\label{3groupings}
\end{figure}

We can see that given large $J$ and $P$, we have to estimate a total of $J^2 P$ parameters, which is much larger than the moderate number of observations $JT$, i.e. $JP \gg T$. Consequently, ordinary least squares estimation is not feasible. Additionally, due to the structural change points in the macro and financial data (although not explored in this paper), the effective number of observations used for estimation could be much smaller than the original $T$. Thus we can see that on one hand, we do not want to impose any restrictions on the parameters and attain some general representations; on the other hand, it is known that making the model unnecessarily complex can degrade the efficiency of the resulting parameter estimate and yield less accurate predictions, as well as making interpretation and variable selection difficult. Hence, to avoid over fitting, \textit{regularization} and \textit{variable selection} are necessary. In the following, we are going to discuss the estimation procedure with different kinds of regularization (illustrated in Figure \ref{3groupings}). Before moving on, we incorporate the following very mild belief, as also considered in \cite{RePEc:jae:japmet:v:25:y:2010:i:1:p:71-92}: \textit{the more recent lags should provide more reliable information than the more distant ones}, which tries to strike a balance between attaining model simplicity and keeping the historic information.

\subsection{Universal Grouping} \label{unig}
Without loss of generality, we start from considering one coefficient matrix, say $B_p$ with entries $\{B_{pij}, {1 \leqslant i,j \leqslant J}\}$. Inspired by \cite{RePEc:jae:japmet:v:25:y:2010:i:1:p:71-92}, we note the fact that the dynamic of some variable is driven by itself, while for a different variable, it might be driven by the dynamics of others. Consequently, we treat the variables' own lags (diagonal terms of $B_p$) different from others' lags (off-diagonal terms of $B_p$) and impose different regularizations for them. We assume that the off-diagonal coefficients of $B_p$ are not only sparse, but also have the same sparsity pattern across different columns, which we call \textit{group sparsity}. Thus we base our selection solution on group Lasso techniques (\cite{RePEc:bla:jorssb:v:68:y:2006:i:1:p:49-67}) for the off-diagonal terms and Lasso techniques (\cite{ti96}) for the diagonal terms here. We use $B_{pj-j}$ to denote the vector composed of $\{B_{pji}\}_{i \neq j}$ and $W_{-j}$ to denote the $(J-1) \times (J-1)$ diagonal matrix $\diag[w_1, \ldots, w_{j-1}, w_{j+1}, \ldots, w_J]$ where $w_i$ is the positive real-valued weight associated with the $i$th variable for $1 \leqslant i \leqslant J$. It is included here primarily for practical implementation since if $w_i$ is chosen as the $\std(Y_i)$, it is equivalent (subsection \ref{algorithm} for details) to standardize the predictors so that they all have zero mean and unit variance, \cite{ti96}, which is also preferable for comparisons to prior works.

Specifically, given the above notations, we use the group Lasso type penalty $\sum_{j=1}^J \|B_{pj-j} W_{-j} \|_2$ and Lasso type penalty $\mu \sum_{j=1}^J w_j|B_{pjj}|$ to impose regularizations on other regressors' lags and predicted variables' own lags respectively and have the following penalty for the $B_p$ matrix:
\begin{equation}
\sum_{j=1}^J \|B_{pj-j} W_{-j} \|_2 + \mu \sum_{j=1}^J w_j|B_{pjj}|  \leqslant C p^{-\alpha}, \label{bppen}
\end{equation}
with some generic constant $C$.
\begin{itemize}
\item The hyperparameter $\mu$ controls the extent to which others' lags are less (more) ``important'' than the own lags. When $\mu$ is large, the penalty assigned to own lags is larger than to others' lags. As a result, it is more likely that the off-diagonal entries are shrunk to $0$ instead of the diagonal ones, which corresponds to the case that the variable's dynamic is driven by itself, and vice versa when $\mu$ is small.
\item The item $p^{-\alpha}$ reflects different regularization for different lags (over time). It becomes smaller when $p$ gets larger. This is consistent with the previous belief: \textit{the more recent lags should provide more reliable information than the more distant ones}. Thus as a result, large amounts of shrinkage are towards the more distant lags, whereas small amounts of shrinkage are towards the more recent ones. The hyperparameter $\bf \alpha$ governs the relative importance of distant lags w.r.t. the more recent ones. Other decreasing functions of $p$, e.g. $f(p)=\log(p)^{-\alpha}, f(p)=\exp(p)^{-\alpha}$ could also be used. However, we do not consider a general representation (and use a data driven way to estimate $f(1), \ldots, f(p), \ldots, f(P)$ correspondingly) to avoid too many tuning parameters, especially when $P \rightarrow \infty$.
\end{itemize}
 Since we have $P$ coefficient matrices $B_1, \ldots, B_P$, summing \eqref{bppen} up over $p$ (after multiplying $p^\alpha$ on both sides) yields $\sum_{p=1}^P \sum_{j=1}^J p^{\alpha} \|B_{pj-j} W_{-j} \|_2 + \mu \sum_{p=1}^P \sum_{j=1}^J p^{\alpha} w_j|B_{pjj}|  \leqslant C P$. If we couple this to the quadratic loss $\{2J(T-P)\}^{-1} \sum_{t=P+1}^T \|Y_t^\T-X_t^\T B\|_2^2$ through Lagrange multipliers, we have equation \eqref{eq:unig}:
\begin{eqnarray}
&& \min_B\{J(T-P)\}^{-1} \sum_{t=P+1}^T \|Y_t^\T-X_t^\T B\|_2^2 +  \lambda \Big(\sum_{p=1}^P \sum_{j=1}^J p^{\alpha} \|B_{pj-j} W_{-j} \|_2 + \mu \sum_{p=1}^P \sum_{j=1}^J p^{\alpha} w_j|B_{pjj}|\Big)\nonumber \\
&\stackrel{\gamma=\lambda\mu}{=}&\min_B\{J(T-P)\}^{-1} \sum_{t=P+1}^T \|Y_t^\T-X_t^\T B\|_2^2 +  \lambda \sum_{p=1}^P \sum_{j=1}^J p^{\alpha} \|B_{pj-j} W_{-j} \|_2 + \gamma \sum_{p=1}^P \sum_{j=1}^J p^{\alpha} w_j|B_{pjj}|\label{eq:unig}
\end{eqnarray}
with hyperparameters $\lambda$, $\gamma$ and $\alpha$. We call this estimate $\hat B$ the {\bf \textit{universal grouping}} estimate. As the number of variables $J$ increases, the autocoefficients should be shrunk more in order to avoid over-fitting, as already discussed by \cite{DeMol2008318}.

 Using the group Lasso type regularization for the off-diagonal terms actually poses some strong assumptions on the underlying structure, which is not realistic from an economic point of view. \textbf{Remark 2.2.1} {\textit{First}}, we just have one hyperparameter $\mu$ ($\mu =\gamma/\lambda$) to control the relative weights between own lags and others' lags. This means that the weights between own's lags and others' lags are the same across different dimensions which is hardly met in practice. Correspondingly, when we select the ``optimal'' $\mu$ to optimize the forecasting performance, we are actually optimizing the averaged forecasting performance for all $J$ variables instead of the variable of particular interest. This might produce suboptimal forecasts. \cite{RePEc:jae:japmet:v:25:y:2010:i:1:p:71-92} considers a special case that own lags are always more ``important'' than others' lags, which might be less general than ours. \textbf{Remark 2.2.2} {\textit{Second}}, using the $L_2$ norm $\|B_{pj-j} W_{-j} \|_2$ might shrink all off-diagonal terms in the same row ($\{B_{pj\cdot}\}_{\cdot \neq j}$) to zero simultaneously, which implicitly means that, for the $j$th corresponding variable, we assume it is either significant for all the other $J-1$ variables or not for any other $J-1$ variables at all. This is, again, too strong from an economic point of view.

\subsection{No Grouping} \label{nog}
To amend the deficiencies of the universal grouping estimate, we estimate the autocoefficient matrix $B$ column by column instead of all at once. Without loss of generality, we consider the $j$th column $B_{\cdot \cdot j}$ here. Since $B_{\cdot \cdot j}$ is a vector, we can use the Lasso type penalties for both own lags and others' lags. By following similar ideas and abbreviations in subsection \ref{unig}, we have equation \eqref{nog1} to get ${\hat B_{\cdot \cdot j}}$:
\begin{eqnarray}
&& \min_{B_{\cdot \cdot j}}(T-P)^{-1} \sum_{t=P+1}^T (Y_{tj}-X_t^\T B_{\cdot \cdot j})^2 +  \lambda_j \Big(\sum_{p=1}^{P} \sum_{i \neq j} p^\alpha w_i |B_{pij}| +  u_j {\sum_{p=1}^{P} p^\alpha w_j |B_{pjj}|}\Big)\nonumber\\
&\stackrel{\gamma_j=\lambda_j\mu_j}{=}&\min_{B_{\cdot \cdot j}}(T-P)^{-1} \sum_{t=P+1}^T (Y_{tj}-X_t^\T B_{\cdot \cdot j})^2 +  \lambda_j \sum_{p=1}^{P} \sum_{i \neq j} p^\alpha w_i |B_{pij}| +  \gamma_j {\sum_{p=1}^{P} p^\alpha w_j |B_{pjj}|}\label{nog1}
\end{eqnarray}
with hyperparameters $\lambda_j$, $\gamma_j$ and $\alpha$. The subindex $j$ is added to $\lambda_j$ and $\gamma_j$ to emphasize that they could vary when estimating different ${B_{\cdot \cdot j}}$'s, $1\leqslant j \leqslant J$. We call this estimate $\hat B = (\hat B_{\cdot \cdot 1}, \ldots, \hat B_{\cdot \cdot J})$ the {\bf \textit{no grouping}} estimate.

\textbf{Remark 2.3.1} Because of different $\mu_j$'s ($\mu_j =\gamma_j/\lambda_j$) for different columns' estimates $\hat B_{\cdot \cdot j}$, we allow individualized weights between own lags and others' lags and could tune $\lambda_j$'s and $\gamma_j$'s to produce optimal forecasting performance for each variable of interest, say the $j$th. \textbf{Remark 2.3.2} Also for the same reason, we could get rid of the disadvantage that all off-diagonal terms in one row might be shrunk to $0$ simultaneously.

For simplicity of notation, we drop the common subindex $j$ and write $Y_{tj} = y_t$, $B_{\cdot \cdot j} = \beta$, $B_{pij} = c_i, i \neq j$, $B_{pjj} = d_p$, $\lambda_j = \lambda$, $\gamma_j=\gamma$, and \eqref{nog1} becomes:
\begin{eqnarray}
\min_{\beta} Q_T(\beta)&=& \min_\beta(T-P)^{-1} \sum_{t=P+1}^T (y_t-X_t^\T \beta)^2 +  \lambda \sum_{p=1}^{P} \sum_{i \neq j} p^\alpha w_i |c_i| +  \gamma {\sum_{p=1}^{P} p^\alpha w_p |d_p|} \nonumber \\ &=& \min_\beta (T-P)^{-1} \sum_{t=P+1}^T (y_t-X_t^\T \beta)^2+  \lambda_{i}\sum_{i=1}^{P(J-1)} |c_{i}| +  {\gamma_{p}\sum_{p=1}^{P} |d_{p}|}\label{nog2}
\end{eqnarray}
with $\lambda_i=\lambda p^\alpha w_i$ and $\gamma_p=\gamma p^{\alpha} w_p$.

\subsection{Segmentized Grouping} \label{segg}
Since the large panel of macroeconomic and financial data sets usually have some natural ``segment'' structure, e.g. multiple interest rate time series w.r.t. different maturities, different number of employees w.r.t. different industrial sectors, different price indices w.r.t. different goods etc, if we take this information into account instead of estimating $B$ either all at once or column by column, we could also do it segment by segment. Without loss of generality, we consider the $i$th segment $B_{\cdot \cdot \mathcal N_i}$. $\mathcal N_i$ is the index set for the $i$th segment and $N_i = |\mathcal N_i|$ denotes the cardinality of the set $\mathcal N_i$. $Y^\T_{t\mathcal N_i}$ is the corresponding part of $Y_t^\T$. We also use $W_{\mathcal N_i}$ to denote the $N_i \times N_i$ diagonal matrix with diagonal entries $\{w_i\}_{i \in {\mathcal N_i}}$ and $W_{\mathcal N_i-j}$ to denote the $(N_i-1) \times (N_i-1)$ diagonal matrix with diagonal entries $\{w_i\}_{i \in {\mathcal N_i, i \neq j}}$.

Under this situation, we have: own lags, others' (in the same segment) lags and others' (outside the segment) lags for the estimation of the $i$th segment's corresponding autoregressive coefficients $B_{\cdot \cdot \mathcal N_i}$. Also following similar ideas and abbreviations in subsection \ref{unig}, we have the following estimation equation:
\begin{eqnarray}
&\min_{B_{\cdot \cdot N_i}}\{N_i(T-P)\}^{-1} \sum_{t=P+1}^T \|Y_{t\mathcal N_i}^\T-X_t^\T B_{\cdot \cdot \mathcal N_i}\|_2^2 \nonumber \\&+  \lambda_{\mathcal N_i} \Big(\sum_{p=1}^{P}  \sum_{j \notin N_i} p^{\alpha} \|B_{pj\cdot}W_{\mathcal N_i}\|_2 + { \mu_{1\mathcal N_i} \sum_{p=1}^{P} p^{\alpha} w_j |B_{pjj}|} +  {\mu_{2\mathcal N_i}\sum_{p=1}^{P} \sum_{j \in N_i} p^{\alpha} \|B_{pj-j}W_{\mathcal N_i-j}\|_2}\Big)\nonumber \\
&\min_{B_{\cdot \cdot N_i}}\{N_i(T-P)\}^{-1} \sum_{t=P+1}^T \|Y_{t\mathcal N_i}^\T-X_t^\T B_{\cdot \cdot \mathcal N_i}\|_2^2 \nonumber \\&+  \lambda_{\mathcal N_i} \sum_{p=1}^{P}  \sum_{j \notin \mathcal N_i} p^{\alpha} \|B_{pj\cdot}W_{\mathcal N_i}\|_2 + { \gamma_{\mathcal N_i} \sum_{p=1}^{P} p^{\alpha} w_j |B_{pjj}|} +  {\eta_{\mathcal N_i}\sum_{p=1}^{P} \sum_{j \in N_i} p^{\alpha} \|B_{pj-j}W_{\mathcal N_i-j}\|_2}\label{eq:segg}
\end{eqnarray}
with hyperparameters $\lambda_{\mathcal N_i}, \gamma_{\mathcal N_i}, \eta_{\mathcal N_i}, \alpha$, $\gamma_{\mathcal N_i} = \lambda_{\mathcal N_i} \mu_{1\mathcal N_i}$ and $\eta_{\mathcal N_i} = \lambda_{\mathcal N_i} \mu_{2\mathcal N_i}$, $i=1, \ldots, I$ where $I$ is the overall number of segments. We call this estimate $\hat B = (\hat B_{\cdot \cdot \mathcal N_1}, \ldots, \hat B_{\cdot \cdot \mathcal N_I})$ the {\bf \textit{segmentized grouping}} estimate.

\cite{ch:pe:07} consider the case $P=1$ and $T$ is large (relative to $J$) through some ``neighboring'' procedure, which can be viewed as a special case of the ``segmentized grouping'' we studied here.

\subsection{Forecast Evaluation and Choice of Parameters} \label{algorithm2}
\begin{figure}
\centering
  \includegraphics[width=13cm]{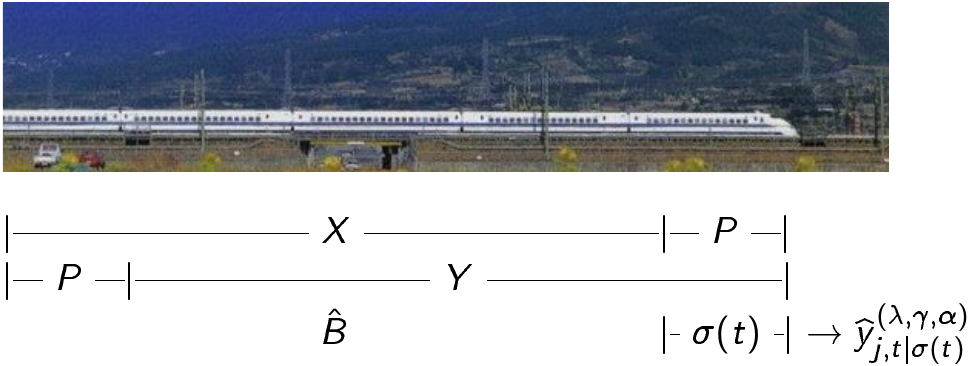}
\caption{Illustration of the Rolling Scheme}   \label{rolling}
\end{figure}

The three penalization methods discussed above critically depend on penalty parameter selection for their performance in model selection, parameter estimation and prediction accuracy. Here we have hyper-parameters $\lambda, \gamma$ (universal grouping), $\lambda_j, \gamma_j, 1\leqslant j \leqslant J$ (universal grouping), $\lambda_i, \gamma_i, \eta_i, 1\leqslant i \leqslant I$ and $\alpha$, and choose them via a data driven ``rolling scheme''. To simulate real-time forecasting, we conduct an out-of-sample experiment. Let $T_0$ and $T_1$ denote the beginning and the end of the evaluation sample respectively. The point estimate of the $j$th variable's forecast is denoted by $\widehat y_{j, t|\sigma(t)}^{(\lambda, \gamma, \alpha)}$ based on $\sigma(t)$, the information up to time $t$. The point estimate of the one-step-ahead forecast is computed as in equation \eqref{nog1}, and the $h$-step-ahead forecasts are computed in similar spirit. Out-of-sample forecast accuracy is measured in terms of mean squared forecast error (MSFE):
$$MSFE_{j,h}^{(\lambda, \gamma, \alpha)} = \frac{1}{ T_1-T_0-h+1 }\sum_{t=T_0}^{T_1-h}(\widehat y_{j, t+h|\sigma(t)}^{(\lambda, \gamma, \alpha)}- y_{j, t+h|\sigma(t)})^{2}.$$
We report results for MSFE relative to the benchmark (random walk with drift) model's (abbreviated as ${MSFE_{j,h}^{(0)}}$), as also considered by \cite{RePEc:jae:japmet:v:25:y:2010:i:1:p:71-92}, i.e.
$$RMSFE_{j,h}^{(\lambda, \gamma, \alpha)}= \frac{MSFE_{j,h}^{(\lambda, \gamma, \alpha)}}{MSFE_{j,h}^{(0)}}.$$
The parameters are estimated using the observations from the most recent $10$ years (rolling scheme) as illustrated in Figure \ref{rolling}. The parameters are set to yield a desired fit for the variable(s) of interest from $T_0$ to $T_1$. In other words, to obtain the desired magnitude of fit, the search is performed over a grid of $\lambda, \gamma$ and $\alpha$ to minimize $\sum_{j=1}^J RMSFE_{j,h}^{(\lambda, \gamma, \alpha)}$ (universal grouping); $\lambda_j, \gamma_j$ and $\alpha$ to minimize $RMSFE_{j,h}^{(\lambda, \gamma, \alpha)}$ (no grouping); $\lambda_i, \gamma_i, \eta_i$ and $\alpha$ to minimize $\sum_{j \in \mathcal{N}_i} RMSFE_{j,h}^{(\lambda, \gamma, \eta, \alpha)}$ (segmentized grouping) respectively. Due to computational cost, we prefix $\alpha$ to be $1$ or $2$ first, and then do the search of $\lambda$'s and $\gamma$'s over loose grids. For the nice performing $\lambda$'s and $\gamma$'s, we search over denser grids around them afterwards. The \textit{parfor} command in Matlab is used to facilitate parallel computations to fasten this process. Also, using the least angle regression package provided at www-stat.stanford.edu/$\sim$tibs/glmnet-matlab, makes the computation time together with the parameter selection very moderate in our experience.

\subsection{Algorithm} \label{algorithm}
Motivated by the adaptive lasso procedure, \cite{RePEc:bes:jnlasa:v:101:y:2006:p:1418-1429}, if we define
\begin{itemize}
\item $\mathcal{P}=\diag[1^{\alpha}, 2^{\alpha}, \ldots, P^{\alpha}] \otimes I_{J \times J}$, where $\diag[1^{\alpha}, 2^{\alpha}, \ldots, P^{\alpha}]$ is the diagonal matrix with diagonal entries $\{1^{-\alpha}, 2^{-\alpha}, \ldots, P^{-\alpha}\}$, $\otimes$ is the Kronecker product and $I_{J \times J}$ is the $J \times J$ identity matrix;
    \item $\mathcal{W}=I_{P \times P} \otimes \diag[w_1, w_2, \ldots, w_J]$;
    \item $\tilde X^\T = X^\T \mathcal{W}^{-1} \mathcal{P}^{-1}$ and $\tilde B = \mathcal{P} \mathcal{W}B$
    \end{itemize}
and note the fact that $X^\T B$ in \eqref{eq:main} is the same as $X^\T \mathcal{W}^{-1} \mathcal{P}^{-1} \mathcal{P} \mathcal{W}B= \tilde X^\T \tilde B$, we have the following estimation procedure (the proof is very simple and hence is omitted):
\begin{itemize}
\item[(1)] Generate $\tilde X^\T = X^\T \mathcal{W}^{-1}\mathcal{P}^{-1}$;
\item[(2)] Corresponding to the three different estimates \eqref{eq:unig}, \eqref{nog1} and \eqref{eq:segg}, solve:
\begin{align}
&\min_{\tilde B}\{J(T-P)\}^{-1} \sum_{t=P+1}^T \|Y_t^\T-\tilde X_t^\T \tilde B\|_2^2 +  \lambda \sum_{p=1}^P \sum_{j=1}^J \|\tilde B_{pj-j}\|_2 + \gamma \sum_{p=1}^P \sum_{j=1}^J |\tilde B_{pjj}|,\label{eq:tilunig}\\
&\min_{\tilde B_{\cdot \cdot j}}(T-P)^{-1} \sum_{t=P+1}^T (Y_{tj}-\tilde X_t^\T \tilde B_{\cdot \cdot j})^2 +  \lambda_j \sum_{p=1}^{P} \sum_{i \neq j} |\tilde B_{pij}| +  \gamma_j {\sum_{p=1}^{P} |\tilde B_{pjj}|},\label{tilnog1}\\
&\min_{B_{\cdot \cdot \mathcal N_i}}\{N_i(T-P)\}^{-1} \sum_{t=P+1}^T \|Y_{t\mathcal N_i}^\T-X_t^\T \tilde B_{\cdot \cdot \mathcal N_i}\|_2^2 \nonumber +  \lambda_{\mathcal N_i} \sum_{p=1}^{P}  \sum_{j \notin N_i} \|\tilde B_{pj\cdot}\|_2 + { \gamma_{\mathcal N_i} \sum_{p=1}^{P} |\tilde B_{pjj}|} \nonumber \\ &+  {\eta_{\mathcal N_i}\sum_{p=1}^{P} \sum_{j \in N_i} \|\tilde B_{pj-j}\|_2};\label{eq:tilsegg};
\end{align}
\item[(3)] Output $\hat B = \mathcal{W}^{-1}\mathcal{P}^{-1} {\hat {\tilde B}}$ with $\hat {\tilde B}$ minimizing \eqref{eq:tilunig} (universal grouping),  $\hat {\tilde B} = (\hat {\tilde B}_{\cdot \cdot 1}, \ldots, \hat {\tilde B}_{\cdot \cdot J})$, $\hat {\tilde B}_{\cdot \cdot j}$ minimizing \eqref{tilnog1} (no grouping); $\hat {\tilde B} = (\hat {\tilde B}_{\cdot \cdot \mathcal N_1}, \ldots, \hat {\tilde B}_{\cdot \cdot \mathcal N_I})$, $\hat {\tilde B}_{\cdot \cdot \mathcal N_i}$ minimizing \eqref{eq:tilsegg} (segmentized grouping).
\end{itemize}
At Step (2), motivated by \cite{RePEc:bla:jorssb:v:69:y:2007:i:1:p:63-78}, as we have more than one penalty terms (mixed Lasso and group Lasso), we could iterate between penalties to solve it as the standard (group) Lasso problem.

For the ``no grouping'' estimate, by noting that $\gamma_{j}=\mu_j \lambda_j$ and $\gamma_j {\sum_{p=1}^{P} |\tilde B_{pjj}|} = \lambda_j {\sum_{p=1}^{P} |\mu_j \tilde B_{pjj}|} $, the estimation procedure above is equivalent to:
\begin{itemize}
\item[(1)] Generate $\tilde X^\T = X^\T \mathcal{W'}^{-1}\mathcal{P}^{-1}$ with $\mathcal{W'}=I_{P \times P} \otimes \diag[w_1, w_2, w_{j-1}, u_j w_j, w_{j+1}, \ldots, w_J]$ for estimating $B_{\cdot \cdot j}, 1\leqslant j \leqslant J$;
\item[(2)] Corresponding to \eqref{tilnog1}, solve:
\begin{align}
\min_{\tilde B_{\cdot \cdot j}}(T-P)^{-1} \sum_{t=P+1}^T (Y_{tj}-\tilde X_t^\T \tilde B_{\cdot \cdot j})^2 +  \lambda_j \sum_{p=1}^{P} \sum_{i=1}^J |\tilde B_{pij}|;\label{tilnog11}
\end{align}
\item[(3)] Output $\hat B = \mathcal{W'}^{-1}\mathcal{P}^{-1} {\hat {\tilde B}}$ with $\hat {\tilde B} = (\hat {\tilde B}_{\cdot \cdot 1}, \ldots, \hat {\tilde B}_{\cdot \cdot J})$, $\hat {\tilde B}_{\cdot \cdot j}$ minimizing \eqref{tilnog1} (no grouping).
\end{itemize}
At Step (2), we could avoid iterating between multiple penalties and just solve it as the standard Lasso, e.g. by using the least angle regression package provided at www-stat.stanford.edu/$\sim$tibs/glmnet-matlab.

In ``large $J$, small $T$'' paradigms, to get parsimonious models, shrinkage with penalization in model selection can shrink insignificant regression coefficients towards zero exactly, but at the same time, significant coefficients are shrunk as well though they are retained in selected working models, \cite{Wainwright:2009:STH:1669487.1669506} and \cite{hu:ma:zh:08}. To this end, we only use our method for the variable (and lag) selection, but not for estimation, \cite{ch:be:ha:11}. Thus, we implement the ordinary least squares estimation for the selected variables (and lags) from \eqref{eq:tilunig}, \eqref{tilnog1} and \eqref{eq:tilsegg} w.r.t. three different estimates.

\subsection{Comparison} \label{comparison}
Now it is a matter of what kind of regularization techniques among these three choices to use in practice. {\textit{First}}, as already discussed in Remark 2.2.1 and 2.3.1, from the allowing individualized (for the variable of particular interest) weights between own lags and others' lags and individualized forecasting performance optimization point of view, the ``no grouping'' approach is the best, the ``universal grouping'' one is the worst, and the ``segmentized grouping'' one is in between. {\textit{Second}}, as in Remark 2.2.2 and 2.3.2, from whether all off-diagonal autocoefficients in one row are shrunk to zero point of view, the ``no grouping'' one is still favored. {\textit{Third}}, as in subsection \ref{algorithm2}, the tuning parameters w.r.t. the universal grouping, no grouping or segmentized grouping estimates are selected to optimize the averaged forecasting performance for all variables, the specific variable's forecasting performance or the averaged forecasting performance for the variables in the same segment respectively. When different variables' time series have very distinct patterns, this individualized optimization is preferred. {\textit{Fourth}}, for the estimation of large coefficient matrices, due to the strong \textit{group-sparse} assumption on the underlying structure as mentioned in subsection \ref{unig}, the group lasso type estimator actually has a sharper theoretical risk bound, \cite{hu:zh:09} for more details. In particular, they show that group Lasso is more robust to noise due to the stability associated with group structure and thus requires a smaller sample size to satisfy the sparse eigenvalue condition required in modern sparsity analysis. And the universal grouping estimate is also more computationally efficient since the whole autocoefficient matrix is estimated at once. However, note that the statistical error is a combination of modeling error and estimation error. Even though the group Lasso type estimate might have smaller estimation error, due to the strong assumption to the underlying structure, the overall risk might not be smaller, as we discussed in subsection \ref{unig}. Moreover, the typical macroeconomic data has low frequency, i.e. monthly. Thus the computational cost is not a severe problem since we only need to update the model once per month at most. Due to all these, we suggest the no grouping estimate for practical implementation as a compromise between flexibility and realization of assumptions. For this reason and technical simplicities, we mainly study the theoretical properties of the no grouping estimate as defined in \eqref{nog2} afterwards.

\section{Estimates' Properties}\label{asymptotic}
In this section, we first show that, under the time series setup, if we just use the classic Lasso estimator, the risk bound will depend on the time dependence level as in Theorem \ref{dependent}. To circumvent this problem, through reweighting over time, our estimate in \eqref{nog2} can still produce an estimator, which is shown in Theorem \ref{theorem2} and \ref{theorem3}, to be equivalent to an appropriate oracle. The techniques of the proofs are closely built upon those in \cite{lo09}, \cite{2008arXiv0801.1095B}, \cite{2008arXiv0801.4610L} and \cite{RePEc:bla:jorssb:v:69:y:2007:i:1:p:63-78}.

\subsection{Dependence Matters?}
Now we will illustrate how the temporal dependence level affects the risk bounds of the Lasso type estimator. For technical simplicities, we consider the univariate AR($P$) (or MA($P$)) model with $P \rightarrow \infty $, i.e. $J=1$ for equations \eqref{eq:main0} and \eqref{eq:main}:
 \begin{equation}
 e_t=x_{t1}\theta_1 + \ldots, x_{tP}\theta_P + \epsilon_t = x_t^\T \theta+\epsilon_t, \label{arp}
 \end{equation}
 with the regressors $(x_{t1}, \ldots, x_{tP})=x_t^T$, the coefficients $(\theta_1, \ldots, \theta_P)=\theta^\T$ and the error term $\epsilon_t$. We also define $\bf x$ as a $T \times P$ matrix with the $t,p$th entry as $x_{tp}$ and $e=(e_1, \ldots, e_T)^\T$. $x_{tp}=e_{t-p}$ (or $\epsilon_{t-p}$) corresponds to the AR($P$) (or MA($P$)) model. In this situation, since there are no ``others lags'' ($J=1$) and $\theta$ is a vector, the standard Lasso estimator $\hat \theta$ is defined through:
\begin{eqnarray}
\min_\theta (T-P)^{-1} \sum_{t=P+1}^T (e_t-x_t^\T \theta)^2 +  2\lambda \|\theta\|_1. \label{eq:arp}
\end{eqnarray}
We assume there is a \textit{true} coefficient $\theta^*$ for \eqref{arp} and define $M (\theta^*)=\sum_{p=1}^p \IF (\theta_p^* \neq 0)$ and $M (\hat \theta)=\sum_{p=1}^p \IF (\hat \theta_p \neq 0)$. Before moving on, we recall the fractional cover theory based definition first, which was introduced by \cite{ja:04} and can be viewed as a generalization of $m$-dependency.
Given a set $\mathcal{T}$ and random variables $V_t$, $t\in \mathcal{T}$, we say:
\begin{itemize}
\item A subset $\mathcal{T}'$ of $\ct$ is $independent$ if the corresponding random variables $\{V_t\}_{t\in \ct'}$ are independent.
\item A family $\{\ct_j\}_j$ of subsets of $\ct$ is a $cover$ of $\ct$ if $\bigcup_j \ct_j = \ct$.
\item A family $\{(\ct_j, \bf w_j)\}_j$ of pairs $(\ct_j, \bf w_j)$, where $\ct_j \subseteq \ct$ and $\bf w_j \in [0, 1]$ is a {\it fractional cover} of $\ct$ if $\sum_j \bf w_j \IF_{\ct_j} \geqslant \IF_{\ct}$, i.e. $\sum_{j: t\in \ct_j} \bf w_j \geqslant 1$ for each $t \in \ct$.
\item A (fractional) cover is $proper$ if each set $\ct_j$ in it is independent.
\item $\cx(\ct)$ is the size of the smallest proper cover of $\ct$, i.e. the smallest $m$ such that $\ct$ is the union of $m$ independent subsets.
\item $\cx^*(\ct)$ is the minimum of $\sum_j \bf w_j$ over all proper fractional covers $\{(\ct_j, \bf w_j)\}_j$.
\end{itemize}
Notice that, in spirit of these notations, $\cx(\ct)$ and $\cx^*(\ct)$ depend not only on $\ct$ but also on the family $\{V_t\}_{t \in \ct}$. Further note that $\cx^*(\ct) \geqslant 1$ (unless $\ct =\varnothing$) and that $\cx^*(\ct) =1 $ if and only if the variables $V_t, t\in \ct$ are independent, i.e. $\cx^*(\ct)$ is a measure of the dependence structure of $\{V_t\}_{t \in \ct}$. For example, if $V_t$ only depends on $V_{t-1}, \ldots, V_{t-k}$ but is independent of all $\{V_s\}_{s< t-k}$, we will have $k+1$ independent sets:
\begin{align*}
\ct_1 &=\{V_1, V_{(k+1)+1}, V_{2(k+1)+1}, \ldots\},\\
\ct_2 &=\{V_2, V_{(k+1)+2}, V_{2(k+1)+2}, \ldots\},\\
& \ldots\\
\ct_{k+1} &=\{V_{k+1}, V_{(k+1)+(k+1)}, V_{2(k+1)+(k+1)}, \ldots\},
\end{align*}
s.t. $\bigcup_{j=1}^{k+1} \ct_j = \ct$. So $\cx^*(\ct) =k+1$ (if $k+1 < T$).

Before stating the first main result of this section, we make the following two assumptions.
\begin{assumption}\label{assumption4}
With a high probability $q$, $\forall$ $p$, the random variables $x_{tp}$ and $\epsilon_t$ satisfy
\begin{eqnarray*}
| \epsilon_{t} x_{tp}| \leqslant b_t \; \textrm{and} \quad T^{-1}\summ t1T b_t^2  \leqslant C'
\end{eqnarray*}
for some constants $b_t, C' > 0, t=1, \ldots, T$.
\end{assumption}

\begin{assumption}\label{assumption}
There exists a positive number $\kappa = \kappa(s)$ such that
\begin{eqnarray*}
\min \Big\{ \frac{|x^\T \Delta|_2}{ \sqrt{T}|\Delta_\rr |_2}: |\rr|\leqslant s, \Delta \in \mathbb{R}^{P} \backslash  \{0\}, \parallel \Delta_{\mathcal{R}^c}\parallel_{1} \leqslant 3 \parallel \Delta_{\rr}\parallel_{1}  \Big\} \geqslant \kappa,
\end{eqnarray*}
where $\mathcal{R}^c$ denotes the complement of the set of indices $\rr$, $\Delta_{\rr}$ denotes the vector formed by the coordinates of the vector $\Delta$ w.r.t. the index set $\rr$.
\end{assumption}
Assumption \ref{assumption} is the \textit{restricted eigenvalue} assumption from \cite{2008arXiv0801.1095B}, which is essentially a restriction on the eigenvalues of the Gram matrix $\Psi_T=x^\T x/T$ as a function of sparsity $s$. To see this, recall the definitions of \textit{restricted eigenvalues} and \textit{restricted correlations} in \cite{2008arXiv0801.1095B}:
\begin{align*}
\psi_{\min}(u)&= \min_{z\in \mathbb{R}^P: 1 \leqslant \mathcal{M}(z) \leqslant u}\frac{z^\T \Psi_T z}{|z|_2^2}, \quad 1\leqslant z \leqslant P,\\
\psi_{\max}(u)&= \max_{z\in \mathbb{R}^P: 1 \leqslant \mathcal{M}(z) \leqslant u} \frac{z^\T \Psi_T z}{|z|_2^2},  \quad 1\leqslant z \leqslant P,\\
\psi_{m_1, m_2} &=\max  \Big\{\frac{f_1^\T x_{I_1}^\T x_{I_2} f_2}{T|f_1|_2 |f_2|_2}: I_1 \bigcap I_2 = \varnothing, |I_i| \leqslant m_i, f_i \in \mathbb{R}^{I_i} \backslash \{0\}, i=1, 2\Big\},
\end{align*}
where $|I_i|$ denotes the cardinality of $I_i$ and $x_{I_i}$ is the $T \times |I_i|$ submatrix of $x$ obtained by removing from $x$ the columns that do not correspond to the indices in $I_i$. Lemma 4.1 in \cite{2008arXiv0801.1095B} shows that if the \textit{restricted eigenvalue} of the Gram matrix $\Psi_T$ satisfies $\psi_{\min}(2s) > 3 \psi_{s,2s}$ for some integer $1\leqslant s \leqslant P/2$, Assumption \ref{assumption} holds.

We can now state our first main result.
\begin{theorem}\label{dependent}
Consider the model \eqref{arp} for $P \geqslant 3$, $T \geqslant 1$ and random variables $V_t=\epsilon_t x_{tp}, t \in \ct$. Let the random variables $x_{tp}$ and $\epsilon_t$ satisfy Assumption \ref{assumption4} for any $p$, all diagonal elements of the matrix $x^\T x/T$ euqal to $1$, and $M(\theta^*) \leqslant s$. Furthermore, let $\kappa$ be defined as in Assumption \ref{assumption}, and $\phi_{max}$ be the maximum eigenvalue of the matrix $X^\T X/T$. Let $\lambda=\sqrt{{\cx^*(\ct)}{(\log P)^{1+\delta'}C'}/T}, \quad \delta'>0.$
Then with probability at least $q(1- P^{-\delta'})$, for any solution $\hat \theta$ of (\ref{eq:arp}), we have:
\begin{gather}
T^{-1} {\parallel x (\hat \theta - \theta^{*})  \parallel}^2 \leqslant 16 s {\cx^*(\ct)}{(\log P)^{1+\delta'}C'}/{T\kappa^2}, \label{eq:bound1}\\
{\parallel \hat \theta - \theta^{*} \parallel_1} \leqslant 16   \leqslant 16 s \sqrt{{\cx^*(\ct)}{(\log P)^{1+\delta'}C'}/T}/\kappa^2,\label{eq:bound2}\\
M(\hat \theta) \leqslant {64  \phi^2_{max} s}/{\kappa^2}.\label{eq:bound3}
\end{gather}
\end{theorem}
Before explaining the results, we would like to discuss some related results first. Suppose $x$ in \eqref{arp} has full rank $P$ and $\epsilon_t$ is $\N(0, \sigma^2)$. Consider the least squares estimate ($P \leqslant T$) $\hat \theta_{OLS}=(x x^\T)^{-1}x e$. Then from standard least squares theory, we know that the prediction error $\|x^T (\hat \theta_{OLS} - \theta^*)\|_2^2/\sigma^2$ is $\chi_p^2$-distributed, i.e.
\begin{equation}
\E \frac{\|x^T (\hat \theta_{OLS} - \theta^*)\|_2^2}{T}=\frac{\sigma^2}{T}P. \label{olserror}
\end{equation}
In the sparse situation if $\epsilon_t$ is $\N(0, \sigma^2)$ (different from our case), Corollary 6.2 of \cite{bu:va:11} shows that the Lasso estimate obeys the following \textit{oracle inequality}:
\begin{equation}
\frac{\|x^T (\hat \theta_{Lasso} - \theta^*)\|_2^2}{T} \leqslant C_0\frac{\sigma^2 \log P}{T} M (\theta^*) \label{lassoerror}
\end{equation}
with a large probability and some constant $C_0$. The additional $\log P$ factor here could be seen as the price to pay for not knowing the set $\{\theta_p^*, \theta_p^* \neq 0\}$, \cite{do:jo:94}.

Similar to the i.i.d. Gaussian situation discussed above, the term ${s {(\log P)^{1+\delta'}}}$ in \eqref{eq::bound1} could be interpreted as the price to pay for not knowing the set $\{\theta_p^*, \theta_p^* \neq 0\}$. Here we have ${(\log P)^{1+\delta'}}$ instead of $\log P$ because we deviate from the typical i.i.d. Gaussian situation and establish the result under the more general Assumption \ref{assumption4}, which could be thought as the ``finite second moment'' condition. And the $\delta'$ term is the price to pay for this deviation.

For the case of $x_{tp}=\epsilon_{t-p}$ (MA($P$) model), by the definition of $\cx^*(\ct)$ and $V_t$, if $k^*=\max\{p, \textrm{s.t.} \theta^*_p \neq 0, \theta^*_{p+1}= \theta^*_{p+2}=\ldots = \theta^*_{P}=0\})$, we have $\cx^*(\ct)=k^*+1$ (if $k^*+1 < T$). The RHS of \eqref{eq:bound1} becomes $16 s {(k^*+1)}{(\log P)^{1+\delta'}}/{T\kappa^2}$. For the case of $x_{tp}=e_{t-p}$ (AR($P$) model), which is equivalent to MA($\infty$) and $\cx^*(\ct) \leqslant T$, the RHS of \eqref{eq:bound1} becomes $16 s {(\log P)^{1+\delta'}}/{\kappa^2}$. Thus $\cx^*(\ct)$ could be interpreted as a measure on how many past lags $V_t$ depends on.
 Additionally, when the time dependence level increases, by the definition of the Gram matrix, $\kappa$ will decrease since it characterizes how strong $x_{t1}, \ldots, x_{tP}$ depend on each other. Still using the MA($k^*$) example considered above, $\kappa$ could be thought as a measure on how strong $e_{t}$ depends on $e_{t-1}, \ldots, e_{t-k^*}$, which is a complement of the measure of $\cx^*(\ct)$ on the time dependence level. In both cases (MA($P$) or AR($P$)), Theorem \ref{dependent} states that if we use the standard Lasso estimate directly for the time series, the bounds get larger when the dependence level ($\cx^*(\ct)$) increases and $\kappa$ decreases. In other words, the bound is minimized when $\cx^*(\ct)=1$, which corresponds to the independent situation in the literature.  When $\cx^*(\ct)$ reaches $T$, it will be offset by the $T$ in the denominator. Thus the risk bound does not decrease when $T$ increases. The intuition behind is clear: if the dependence level is strong, then the additional information brought by a ``new'' observation will be effectively less, i.e. the overall information from $\{V_t\}_{t=1}^T$ will be less correspondingly, which will result in increasing estimates' risk bounds. Consequently, we expect the selection not to be stable and to be very sensitive to minor perturbation of the data. In this sense, we do not expect variable selection to provide results that lead to clearer economic
interpretation than principal components or Ridge regression.

\subsection{Consistency of Selection}
To study the oracle properties of the estimator in \eqref{nog2}, we assume that there
is a correct model with the regression and autoregression coefficients $\beta^* = (c^{*\T}, d^{*\T})\T=(c^*_1, \ldots, c^*_{P(J-1)}, d^*_1, \ldots, d^*_P)'$. Furthermore, we assume that there are a total of $p_0 \leqslant P(J-1)$ non-zero
other-lag coefficients and $q_0 \leqslant P$ non-zero own-lag coefficients. For convenience, we
define $S_1 = \{1\leqslant i \leqslant P(J-1), c^*_{i} \neq 0\}$, $\hat S_1 = \{1\leqslant i \leqslant P(J-1), \hat c_{i} \neq 0\}$, $S_2 = \{1\leqslant p \leqslant P, d^*_{p} \neq 0\}$ and $\hat S_2 = \{1\leqslant p \leqslant P, \hat d_{p} \neq 0\}$. Then, the sets $S_1$ and $S_2$ contain the indices of the significant others-lag and own-lag coefficients  respectively, and their complements $S_1^c$ and $S_2^c$ contain the indices of the insignificant coefficients. Next, let $c^*_{S_1}$ denote the ${p_0}\times1$ significant other-lag coefficient vector with $\hat c_{S_1}$ being its associated estimator. Moreover, other related parameters and their
corresponding estimators are analogously defined (e.g. $c^*_{S_1^c}, \hat c_{S_1^c}, d^*_{S_2}, \hat d_{S_2}, d^*_{S_2^c}, \hat d_{S_2^c}$). Finally, let $\beta^*_1 = (c^{*'}_{S_1}, d^{*'}_{S_2})'$ and $\beta^*_2 = (c^{*'}_{S_1^c}, d^{*'}_{S_2^c})'$ with corresponding estimates $\hat \beta_1, \hat \beta_2$. To facilitate the study, we also introduce the notations
\begin{eqnarray*}
a_T &\defeq& \max(\lambda_{i}, \gamma_{p}, i \in S_1, p \in S_2 ),\\
b_T &\defeq& \min(\lambda_{i}, \gamma_{p}, {i} \in S_1^c, {p} \in S_2^c ),
\end{eqnarray*}
where $\lambda_{i}$ and $\gamma_{p}$ are functions of $T$. To investigate the theoretical properties of $\hat \beta$, we introduce the following conditions:
\begin{itemize}
\item[A1] The sequence $\{X_t\}$ is independent of $\eps_t$ ($\eps_t \defeq U_{tj}$);
\item[A2] All roots of polynomial $1-\sum_{p=1}^P d_p^* z_p$ are outside the unit circle;
\item[A3] $\eps_t$ has finite fourth-order moment, i.e. $\E(\eps_t^4) < \infty$;
\item[A4] $\forall j$, $Y_{tj}$ (as components of the covariate $X_t$) is strictly {stationary} and {ergodic} with finte second-order moment (i.e. $\E\|Y_{tj}\|^2 < \infty$).
\end{itemize}
The technical conditions above are typically used to assure the $\sqrt{T}$-consistency and asymptotic normality of the unpenalized least squares estimator.

We also rewrite equation \eqref{nog2} as
\begin{equation}
\min_{\beta} Q_T(\beta)= \min_\beta \sum_{t=P+1}^T (y_t-X_t^\T \beta)^2 +  T \sum_{i=1}^{P(J-1)}\lambda_{i} |c_{i}| +  T {\sum_{p=1}^{P} \gamma_{p} |d_{p}|}\label{nog3}
\end{equation}
by multiplying $2(T-P)$ and writing $T-P$ as $T$ without confusion. Define $\sum_{t=P+1}^T (y_t-X_t^\T \beta)^2$ as $L_T(\beta)$, $x_t^T=(X_{j,t}, X_{2j,t},\ldots, X_{Jj,t})$, the own lags corresponding to the coefficients $c$, and $z_t^\T=X_t^\T\setminus x_t^\T$, others' lags corresponding to the coefficients $d$ respectively. Then $X_t^\T \beta = x_t^\T c + z_t^\T d$.

We first investigate the consistency of the estimator of \eqref{nog2}.

\begin{lemma}
\label{lemma1}
Assume that $a_T=\Co(1)$ as $T \rightarrow \infty$. Then, under conditions (A1-A4), $\exists$ a local minimizer $\widehat \beta$ of \eqref{nog2} s.t.
$$\|\widehat \beta - \beta^*\|=\CO_p(T^{-1/2}+a_T).$$
\end{lemma}
The proof is given in the appendix. Lemma \ref{lemma1} implies that, if the tuning parameters associated with the significant regressors converge to $0$ at a speed faster than $T^{-1/2}$, then there is a local minimizer of \eqref{nog2}, which is $\sqrt{T}$-consistent.
Next, we show that, if the tuning parameters associated with the insignificant regressors shrink to $0$ slower than $T^{-1/2}$, then their coefficients can be estimated exactly as $0$ with probability tending to $1$.

\begin{theorem}[Consistency of Selection] \label{theorem2}
Assume that $b_T \sqrt{T} \rightarrow \infty$ and $\|\widehat \beta - \beta^*\|=\CO(T^{-1/2})$, then
{$$\P(\widehat \beta_2=0)\rightarrow 1.$$}
\end{theorem}
Theorem \ref{theorem2} shows that our method can produce a sparse solution for insignificant coefficients consistently. Furthermore, this
theorem, together with Lemma \ref{lemma1}, indicates that the $\sqrt{T}$-consistent estimator must satisfy $\P(\widehat \beta_2=0)\rightarrow 1$ when the tuning parameters fulfill the appropriate conditions. Finally, we obtain the asymptotic
distribution of this estimator.

\begin{theorem}
\label{theorem3}
Assume that $a_T \sqrt{T} \rightarrow 0$ and $b_T \sqrt{T} \rightarrow \infty$. Then, under conditions (A1-A4), the ``nonzero'' components $\widehat \beta_1$ of the local minimizer $\widehat \beta$  in Lemma \ref{lemma1} satisfies
$$(\widehat \beta_1 - \beta^*_1)\sqrt{T} \stackrel{d}{\rightarrow} \N(0, \Sigma_0^{-1}),$$
$$\P(\hat S_1 = S_1) \rightarrow 1, \quad \P(\hat S_2 = S_2) \rightarrow 1,$$
where $\Sigma_0$ is the submatrix of $\Sigma$ corresponding to $\beta^*_1$, $\Sigma=\diag(B, C)$, $B=\E(x_t x_t^\T)$ and $C=\E(z_t z_t^\T)$.
\end{theorem}
Theorem \ref{theorem3} implies that, if the tuning parameters satisfy the conditions $a_T \sqrt{T} \rightarrow 0$ and $b_T \sqrt{T} \rightarrow \infty$, then, asymptotically, the resulting
estimator can be as efficient as the oracle estimator. And our method can produce a sparse solution for significant coefficients consistently.

Since consistency of selection is established here, if we use the ordinary least squares estimation for the selected variables, we can avoid the $\log$ term on \eqref{lassoerror} and \eqref{eq:bound1}.

\section{Application} \label{application}
We use the dataset of \cite{sto:wat:05a} for illustration. This dataset contains $131$ monthly macro indicators covering a broad range of categories including income, industrial production, capacity, employment and unemployment, consumer prices, producer prices, wages, housing starts, inventories and orders, stock prices, interest rates for different maturities, exchange rates, money aggregates and so on. The time span is from January $1959$ to December $2003$. We apply logarithms to most of the series except those already expressed in rates. The variables of special interest include a measure of real economic activity, a measure of prices and a monetary policy instrument. As in \cite{Ch:Ei:Ev:99}, we use employment as an indicator of real economic activity measured by the number of employees on non-farm payrolls (EMPL). The level of prices is measured by the consumer price index (CPI)
and the monetary policy instrument is the Federal Funds Rate (FFR). All $131$ variables' lags are used as regressors. As discussed earlier, because of the stationary requirement of our method, the series are transformed to obtain stationarity so that many of the series are (2nd order) differences of the raw data series (or logarithm of the raw series).

\begin{figure}
\centering
    \includegraphics[width=14cm]{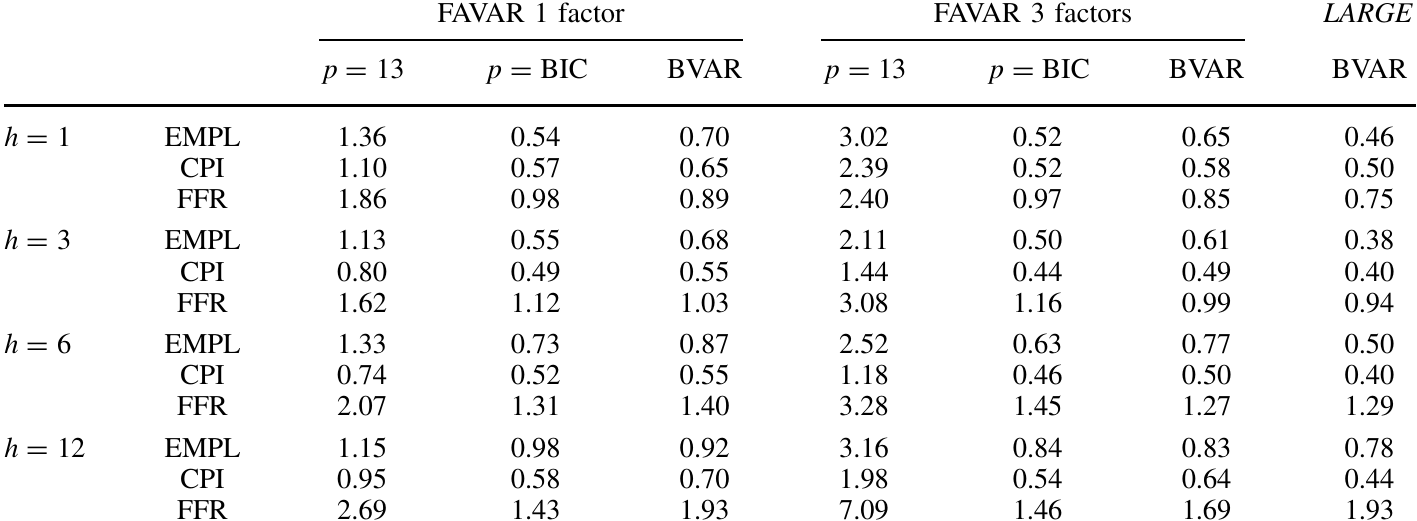}
  \caption{Results of \cite{RePEc:tpr:qjecon:v:120:y:2005:i:1:p:387-422} and \cite{RePEc:jae:japmet:v:25:y:2010:i:1:p:71-92}}\label{bernanke}
\end{figure}

We evaluate the forecast performance over the period from $T_0=$ January $70$ to $T_1=$ December $03$ and for forecast horizons up to one year ($h=1, 3, 6, 12$). The order of the VAR is set to be $P=1, 4, 7, 13, 25$. The resulting performance is summarized in Table \ref{rmsfe} with comparisions to the ones of \cite{RePEc:jae:japmet:v:25:y:2010:i:1:p:71-92} listed under the ``BVAR'' column. As we can see, unlike the information criteria based on lag selection techniques, the RMSFE is very robust to the initial choice of $P$, which primarily benefits from the ``re-weighting over lags'' technique ($p^{-\alpha}$) we used before. For this specific data set, $P=1$ seems enough. But in general, since we never know the true value of lags, we can include a large enough $P$ at the beginning to allow flexibility without worrying about over fitting. Moreover, for the one-step-ahead forecast, our method outperforms for EMPL, CPI and FFR, while when $h \geqslant 3$, it outperforms mainly for EMPL and FFR, especially for the latter one. This results from the fact that different time series might have quite different behaviors, so if we just have the ``universal'' penalty parameter for all of them as in \cite{RePEc:jae:japmet:v:25:y:2010:i:1:p:71-92}, the corresponding forecasting performance might not be optimized. For reference purpose, we also provide the factor-augmented vector autoregressive results of \cite{RePEc:tpr:qjecon:v:120:y:2005:i:1:p:387-422} in Figure \ref{bernanke}.

\begin{table}
  \centering
  \begin{tabular}{ccccccc|c}
    \hline     \hline
     &  & $P=1$ & $P=4$ & $P=7$ & $P=13$ & $P=25$  & BVAR\\ \hline
     & EMPL & $0.3333$ & $0.3336$ & $0.3338$ & $0.3341$ & $0.3335$ &${\color{gray}0.46}$\\
    $h=1$ & CPI & $0.3623$ & $0.3618$ & $0.3613$ & $0.3621$ & $0.3623$ & ${\color{gray}0.50}$\\
     & FFR & $0.4279$ & $0.4281$ & $0.4281$ & $0.4284$ & $0.4287$ & ${\color{gray}0.75}$\\     \hline
          & EMPL & $0.5191$&    $0.5188$&    $0.5192$&    $0.5191$&    $0.5189$& $0.38$ \\
    $h=3$ & CPI & $  0.4990$&    $0.4992$&    $0.4986$&    $0.4995$ &   $0.4996$ & $0.40$\\
     & FFR & $ 0.4615$&    $0.4614$&    $0.4619$&    $0.4617$& $0.4628$ & ${\color{gray}0.94}$\\    \hline
     & EMPL & $0.4730$&    $0.4730$&    $0.4735$&    $0.4729$&   $0.4736$ & ${\color{gray}0.50}$\\
    $h=6$ & CPI & $    0.4880$&    $0.4874$&    $0.4884$&    $0.4885$&    $0.4891$ & $0.40$\\
     & FFR & $    0.5237$&    $0.5242$&    $0.5243$  &  $0.5243$&    $0.5250$& ${\color{gray}1.29}$ \\    \hline
         & EMPL & $     0.4997$ &   $0.4991$&    $0.4992$&    $0.4997$&    $0.5002$ & ${\color{gray}0.78}$\\
    $h=12$ & CPI & $     0.4689$ &   $0.4687$&    $0.4689$&    $0.4694$&    $0.4686$ & $0.44$\\
     & FFR & $    0.4201$ &   $0.4199$&    $0.4201$&    $0.4200$&    $0.4216$ & ${\color{gray}1.93}$\\
    \hline     \hline
  \end{tabular}
  \caption{RMSFE w.r.t. different choices of $h$ and $P$.} \label{rmsfe}
\end{table}

\section{Concluding Remarks and Discussions}\label{discussion}
To summarize, in this article, we first show that under the time series setup, if we still use the classic Lasso type estimator, the risk bound will increase when the time dependence level increases, however, our method could still achieve the consistency of variable selection under such scenario; second, our method is able to do variable selection and lag selection simultaneously, and is rather robust to the initial choice of lags; third, we allow individualized weights between own and others' lags. All these have been confirmed by the real forecasting performance in the previous section and come at a low computational cost.

Some issues we do not explore here include \textit{nonstationarity}, \textit{rank test, cointegration} and \textit{causal test}. For a typical macroeconomic data set, the nonstationarity comes from seasonality, business cycle and economic developments. In spirit of \cite{s2:wh:jr:10}, this motivates us to add a nonstationary component $U \Gamma$ to equation \eqref{eq:main} as below
$$ Y_t=Z_t\Gamma+ X_tB+ U_t,$$
where $Z_t=(Z_1(t), \ldots, Z_R(t))^\T$ contains $R$ basis functions of time consisting of Fourier series with different frequencies and segment by segment ortho-normal polynomials with corresponding $R\times J$ coefficient matrix $\Gamma$, and $X_t, B$ are the same as in equation \eqref{eq:main0}. Studying this extended model deserves further investigation and will be presented in a separate paper. If we want to consider the rank test, cointegration and causal test, what we need for this high dimensional time series is not the ones in the univariate case, but the high dimensional \textit{simultaneous} tests, which might be much more difficult.

\textit{Heteroscedasticity with Cross-section Correlations}.

We consider $\Cov(U_t)=\Sigma$ with nonzero off-diagonal entries in $\Sigma$. Assume that we have a consistent estimate $\hat \Sigma$ for $\Sigma$ (which is another challenging task since $\Sigma$ is a $J \times J$ matrix) with Cholesky decomposition $\hat \Sigma= C^\T C$, where $C$ is an upper triangular matrix with inverse $D$ (which is also an upper triangular matrix). Without loss of generality, assume all diagonal entries of $\hat \Sigma, C$ and $D$ are equal to $1$.
Transform the original $X_t$ by $D$ to generate $\tilde X_t$ ($\tilde X_t=X_tD$) s.t. $\Cov(U_t D)= I$. Under this situation, we are no longer selecting the original variables, but linear transformations of them. Thus we must show that this does not affect the inference. We have
\begin{align*}
&\tilde \beta_1 \tilde  x_{t1} + \tilde  \beta_2 \tilde x_{t2}+ \ldots + \tilde \beta_J \tilde x_{tJ} \\
=& \tilde \beta_1 x_{t1} + \tilde \beta_2 (d_{12}x_{t1} +x_{t2})+ \ldots + \tilde \beta_J (\sum_{j=1}^{J-1} d_{jJ}x_{tj} +x_{tJ})\\
=& (\tilde \beta_1 + \sum_{j=2}^{J} \tilde \beta_j d_{1j})x_{t1} + (\tilde \beta_2 + \sum_{j=3}^{J} \tilde \beta_j d_{2j})x_{t2}+\ldots +       \tilde \beta_J x_{tJ}.
\end{align*}
If the off-diagonal entries of $D$, $\{d_{ij}\}_{i <j}$, are much smaller than the diagonal entries $1$, it is likely that the selected nonzero sets of $\tilde \beta$'s (or $\tilde c, \tilde d$) are the same as the selected nonzero sets $\hat S_1$ and $\hat S_2$ of $\beta$'s, which have been shown to be the same as the \textit{oracle} ones, Theorem \ref{theorem2} and \ref{theorem3}. By the definition of $D$, this means that the off-diagonal entries of $C, \hat \Sigma$ and $\Sigma$ should also be much smaller than their diagonal entries $1$, e.g. the cross-section correlations must be weak enough, which aligns with the case for dynamic factor models, \cite{RePEc:tpr:restat:v:82:y:2000:i:4:p:540-554}.

\vspace{1cm}
{\textbf{Acknowledgement}} The main results of this paper were first presented at the annual Winter Meeting of the Econometric Society, Denver, Jan, $2011$. We are grateful for seminar participants' many interesting comments on several versions of the paper. In particular, I would like to thank Prof Peter Bickel for sponsoring my stay at the University of California, Berkeley. We would also like to thank Prof Marta Ba\'{n}bura, Prof Domenico Giannone and Prof Lucrezia Reichlin for sharing the codes of BVAR.

\section{Appendix}\label{appendix}
\noindent\textbf{Proof of Theorem \ref{dependent}}
The proof of this theorem is based on the ones of Lemma 3.1 and Theorem 3.1 in \cite{lo09} up to a modification of the bound on $\P(\mathcal{A}^c)$ with random event $\mathcal{A}= \Big\{\max_{1\leqslant p \leqslant P} \summ t1T \eps_{t} X_{tp} \leqslant \lambda T \Big\}$, where $n, M$ and $T$ there are equivalent to $T, P$ and $1$ here respectively. The intermediate results in the proof of Theorem 3.1 in \cite{lo09} show that
\begin{gather}
T^{-1} {\parallel X (\widehat B - B^{*})  \parallel}^2 \leqslant 16 s \lambda^2/\kappa^2, \label{eq::bound1}\\
{\parallel \widehat B - B^{*} \parallel} \leqslant 16   \leqslant 16 s \lambda/\kappa^2,\label{eq::bound2}\\
M(\widehat B) \leqslant {64  \phi^2_{max} s}/{\kappa^2}.\label{eq::bound3}
\end{gather}
We have:
\begin{eqnarray*}
\P(\ca^c) = \P\Big(\max_{1\leqslant p \leqslant P} \summ t1T \eps_{t} X_{tp} > \lambda T \Big)
\leqslant P \P \Big(\summ t1T \eps_{t} X_{tp} > \lambda T \Big)
\defeq P \P\Big(\summ t1T V_t > \lambda T\Big).\\
\end{eqnarray*}

Then, by the (extended) Mcdiarmid inequality, see Theorem $2.1$ of \cite{ja:04}, with random vectors $\{V_t\}_{t=1}^T$, we have
\begin{eqnarray*}
\P(\ca^c) \leqslant  P \P\Big(\summ t1T V_t > \lambda T\Big) \leqslant P\exp\left\{-\frac{\lambda^2 T}{\cx^*(\ct) \sum_t b_t^2/T}\right\} \leqslant P^{-\delta'}
\end{eqnarray*}
with $\lambda=\sqrt{{\cx^*(\ct)}{(\log P)^{1+\delta'}C'}/T}, \quad \delta'>0$, which, together with
\eqref{eq::bound1}, \eqref{eq::bound2} and \eqref{eq::bound3}, leads to \eqref{eq:bound1}, \eqref{eq:bound2} and \eqref{eq:bound3}. \hfill $\qquad\square$

\noindent\textbf{Proof of Lemma \ref{lemma1}}
The proofs are closely built upon those of \cite{RePEc:bla:jorssb:v:69:y:2007:i:1:p:63-78}. Let $\delta = (u^\T, v^\T)^\T$, $u=(u_1, \ldots, u_{P(J-1)})^\T$, $v=(v_1, \ldots, v_{P})^\T$, $\alpha_T= T^{-1/2} +a_n$ and $\{\beta^* + \alpha_T \delta: \|\delta\|\leqslant e\}$ be the ball around $\beta^*$. Then, for $\|\delta\|=e$, we have
\begin{align}
D_T (\delta)&= Q_T(\beta^*+\alpha_T \delta) - Q_T(\beta^*) \nonumber\\
&=L_T(\beta^*+\alpha_T \delta) - L_T(\beta^*) + T \sum_{i \in S_1} \lambda_i(|c_i^*+\alpha_T u_i| - \|c_i^*\|)+ T \sum_{j \in S_2} \gamma_j (|d_j^*+\alpha_T v_j|- \|d_j^*\|)\nonumber\\
&=L_T(\beta^*+\alpha_T \delta) - L_T(\beta^*) - T \alpha_T \sum_{i \in S_1} \lambda_i |u_i|- T \alpha_T \sum_{j \in S_2} \gamma_j |v_j|\nonumber\\
&=L_T(\beta^*+\alpha_T \delta) - L_T(\beta^*) - T \alpha_T^2 p_0 e  -T \alpha_T^2 q_0 e \nonumber \\
&=L_T(\beta^*+\alpha_T \delta) - L_T(\beta^*) - T \alpha_T^2 (p_0+q_0) e. \label{eq:10}
\end{align}
Furthermore,
\begin{eqnarray}
L_T(\beta^*+\alpha_T \delta) - L_T(\beta^*) &=& \sum_t \{u_t -a_T z_t^\T v -a_T u^\T x_t\}^2-\sum_t u_t^2 \nonumber\\
&=&a_T^2 \sum_t \{(z_t^\T v)^2 + u^\T x_t x_t^{\T} u \}\label{a1}\\
&&-2a_T \sum_t u_t z_t^\T v \label{a2}\\
&& + 2 a_T^2 \sum_t z_t^\T v u^\T x_t \label{a3}.
\end{eqnarray}
By employing the martingale central limit theorem and the ergodic theorem, we can show that \eqref{a1} $= T a_T^2 \{\delta^\T \Sigma \delta + \Co_p(1)\}$, \eqref{a2} $=\delta^\T \CO_p(T a_T^2)$ and \eqref{a3} $=T a_T^2 \Co_p(1)=\Co_p(T a_T^2)$. Because \eqref{a1} dominates the terms \eqref{a2}, \eqref{a3} and $T \alpha_T^2 (p_0+q_0) e$ in equation \eqref{eq:10}, for any given $\epsilon >0$, there is a large constant $e$ such that
$$\P[\inf_{\|\delta\|=e} Q_T(\beta^*+\alpha_T \delta) > Q_T(\beta^*)]\geqslant 1-\epsilon.$$
This implies that, with probability at least $1-\epsilon$, there is a local minimizer in the ball $\{\beta^* + \alpha_T \delta: \|\delta\|\leqslant e\}$, \cite{bi:kl:ri:we:98} and \cite{Fan_Li_2001}. Consequently, there is a local minimizer of $Q_T(\beta)$ such that $\|\hat \beta-\beta^*\|=\CO_p(\alpha_T)$. This completes the proof. \hfill $\qquad\square$

\noindent\textbf{Proof of Theorem \ref{theorem2}}
The proof is essentially the same as those of Theorem 2 of \cite{RePEc:bes:jnlasa:v:101:y:2006:p:1418-1429} and \cite{RePEc:bla:jorssb:v:69:y:2007:i:1:p:63-78}. For $i \in S_1^c$, assume that there is a local minimizer $\hat \beta$ with $\hat c_i \neq 0$. By the KKT optimality condition, we have
\begin{eqnarray}
0 &=& \frac{\partial L_T(\hat \beta)}{c_i} + T \lambda_i sgn(\hat c_i)\nonumber\\
&=& \frac{\partial L_T(\beta^*)}{c_i} +T \Sigma_i (\hat \beta-\beta^*)\{1+\Co_p(1)\} + T \lambda_i sgn(\hat c_i), \label{eq:12}
\end{eqnarray}
where $\Sigma_i$ denotes the $i$th row of $\Sigma$ and $i \in S_1^c$. By employing the central limit theorem, the first term in equation \eqref{eq:12} is of order $\CO_p(T^{1/2})$. Furthermore, the condition in Theorem \ref{theorem2} implies that its second term is also of order $\CO_p(T^{1/2})$. Both are dominated by $T \lambda_i$ since $b_T \sqrt{T} \rightarrow \infty$. Therefore, the sign of equation \eqref{eq:12} is dominated by the sign of $\hat c_i$.
Thus \eqref{eq:12} can not be equal to $0$. Consequently, we must have $\hat c_i =0$ in probability. Analogously, we can show that $\P(\hat d_{S_2^c}=0)\rightarrow 1$. This completes the proof. \hfill $\qquad\square$

\noindent\textbf{Proof of Theorem \ref{theorem3}}
Applying Lemma \ref{lemma1} and Theorem \ref{theorem2}, we have {$\P(\widehat \beta_2=0)\rightarrow 1$}. Hence, the minimizer of $Q_T(\beta)$ is the same as that of $Q_T(\beta_1)$ with probability tending to $1$. This implies that the estimator $\hat \beta_1$ satisfies the equation
\begin{equation}
\frac{\partial Q_T(\beta_1)}{\beta_1}\Big|_{\beta_1=\hat \beta_1}=0. \label{eq:13}
\end{equation}
According to Lemma \ref{lemma1}, $\hat \beta_1$ is $\sqrt{T}$-consistent. Thus, the Taylor series expansion of equation \eqref{eq:13} yields
\begin{eqnarray*}
0 &=& \frac{1}{\sqrt{T}} \frac{\partial L_T(\hat \beta_1)}{\beta_1} + g(\hat \beta_1) \sqrt{T}\\
&=&\frac{1}{\sqrt{T}} \frac{\partial L_T(\beta^*_1)}{\beta_1} + g(\beta^*_1) \sqrt{T} + \Sigma_0 \sqrt{T} (\hat \beta_1-\beta_1^*)+\Co_p(1),
\end{eqnarray*}
where $g$ is the first-order derivative of the penalty function
$$ \sum_{i \in S_1} \lambda_i |c_i| + \sum_{j \in S_2} \gamma_j |d_j|,$$
and $g(\hat \beta_1)=g(\beta_1^*)$ when $T$ is sufficiently large. Furthermore, it can be easily shown that $g(\beta_1^*)\sqrt{T}=\Co_p(1)$, which implies that
\begin{eqnarray*}
(\hat \beta_1 -\beta_1^*) \sqrt{T} &=& \frac{\Sigma_0^{-1}}{\sqrt{T}} \frac{\partial L_T(\beta_1^*)}{\partial \beta_1} + \Co_p(1)\\
&\stackrel{d}{\rightarrow}& \N(0, \Sigma_0^{-1}).
\end{eqnarray*}
The next step is to show $\P(\hat S_1 = S_1) \rightarrow 1$ and $\P(\hat S_2 = S_2) \rightarrow 1$. $\forall i \in S_1$ and $p \in S_2$, the asymptotic normality result indicates that $\hat c_i \stackrel{p}{\rightarrow} c_i^*$ and $\hat d_p \stackrel{p}{\rightarrow} d_p^*$, where $\stackrel{p}{\rightarrow}$ stands for convergence in probability. Thus $\P(i \in \hat S_1) \rightarrow 1$ and $\P(p \in \hat S_2) \rightarrow 1$. It suffices to show that $\forall i' \notin S_1$ and $p' \notin S_2$, $\P(i' \in \hat S_1) \rightarrow 0$ and $\P(p' \in \hat S_2) \rightarrow 0$, which have been shown by Theorem \ref{theorem2}. This completes the proof. \hfill $\qquad\square$

\bibliography{biball}
\bibliographystyle{apalike}
\end{document}

%% file: defs.tex
\newcommand{\xy}{\{(X_i, Y_i)\}_{i=1}^n}
\newcommand{\lnh}{l_{n,h}}
\newcommand{\lng}{l_{n,g}}
\def\t{^{\scriptstyle{\mathsf{T}}} \hspace{-0.15em}}
\def\T{\top}
\def\ct{\mathcal{T}}
\def\cx{\mathcal{X}}
\def\ca{\mathcal{A}}
\def\ztt{Z_t^\T}
\def\tztt{\widetilde Z_t^\T}
\def\sfa{\scalefont{0.6}}
\def\jt1{T^{-1}}
\def\tT{1 \leqslant t \leqslant T}
\def\ta{\widetilde A}
\def\sfa{\scalefont{0.6}}
\def\sfb{\scalefont{1.6666666666}}
\def\del{\delta}
\def\al{\alpha}
\def\ti{\tilde}
\def\bb#1{\mathbb{#1}}
     \def\bbs{\ensuremath{{\bb S}}}
\def\t{^{\scriptstyle{\mathsf{T}}} \hspace{-0.15em}}

\newcommand{\eqsplit}[2][*]%
  {\ifthenelse{\equal{#1}{*} \or\equal{#1}{<???>}
                     \or\equal{#1}{nn} \or\equal{#1}{ } \or\equal{#1}{}}
    { \begin{align*}%
         #2 %
         \end{align*}%
        }
    {\begin{equation}\label{#1}\begin{split}\allowdisplaybreaks%
         #2%
         \end{split}\end{equation}
        }
  }

\newcommand{\nit}{\mathbb{N}}
\def\N{\mbox{N}}            
\renewcommand{\i}{\mbox{\bf i}}
\newcommand{\CO}{{\mathcal{O}}}
\def\Co{{\scriptstyle \mathcal{O}}} 
\renewcommand{\O}{\CO}
\renewcommand{\o}{\Co}
\newcommand{\zit}{\mathbb{Z}}
\newcommand{\R}{\mathbb{R}}
\newcommand{\rr}{\mathcal{R}}
\newcommand{\tr}{\mathop{\rm{tr}}}
\newcommand{\ii}{\mathcal{I}}
\newcommand{\diag}{\mathop{\rm{diag}}}
\newcommand{\rank}{\mathop{\rm{rank}}}
\newcommand{\Var}{\mathop{\mbox{\sf Var}}}
\newcommand{\var}{\mathop{\mbox{\sf Var}}}
\newcommand{\Cov}{\mathop{\mbox{\sf Cov}}}
\newcommand{\cov}{\mathop{\mbox{\sf Cov}}}
\newcommand{\q}{\mathop{\mbox{Q}}}
\newcommand{\E}{\mathop{\mbox{\sf E}}}     
\renewcommand{\P}{\operatorname{P}}            
\newcommand{\Corr}{\mathop{\mbox{Corr}}}
\def\std{\mathop{\mbox{Std}}}            
\newcommand{\IF}{\boldsymbol{1}}    
\def\defeq{\stackrel{\mathrm{def}}{=}}  
\def\eqdef{\stackrel{\rm def}{=}}
\def\eps{\ensuremath{\varepsilon}}
\def\iid{i.i.d.\xspace }
\def\argmax{\mathop{\mbox{arg\,max}}}
\def\argmin{\mathop{\mbox{arg\,min}}}
\def\ind{\IF}
\def\en{\infty}
\def\summ#1#2#3{\sum_{#1=#2}^{#3}}
\def\noframe{}
\def\t{^{\scriptstyle{\mathsf{T}}} \hspace{-0.15em}}
\newcommand{\scz}{{\cal Z}}

\newcommand{\alze}{{\alpha^{(0)}}}
\newcommand{\lleq}{\leqslant}
\newcommand{\ggeq}{\geqslant}
\newcommand{\talze}{{\widetilde{\alpha}^{(0)}}}
\newcommand{\zze}{{Z^{(0)}}}
\newcommand{\tzze}{{\widetilde{Z}^{(0)}}}
\newcommand{\hzt}{{\widehat{Z}_t}}
\newcommand{\zzet}{{Z_t^{(0)}}}
\newcommand{\tzzet}{{\widetilde{Z}_t^{(0)}}}
\newcommand{\zzetr}{{Z_t^{(0)\top}}}
\newcommand{\tzzetr}{{\widetilde{Z}_t^{(0)\top}}}

\newcommand{\hZ}{{\widehat{Z}}}
\newcommand{\hA}{{\widehat{A}}}
\newcommand{\ha}{{\widehat{\alpha}}}
\newcommand{\czzetr}{{\cZ_t^{(0)\top}}}
\newcommand{\chA}{{\widehat{\cA}}}
\newcommand{\chZ}{{\widehat{\cZ}}}
\newcommand{\cZ}{{\cal Z}}
\newcommand{\cA}{{\cal A}} 